\documentclass[acmsmall]{acmart}
\setcopyright{none}
\acmDOI{}

\usepackage{amsmath}
\usepackage{amsfonts}
\usepackage{graphicx}
\usepackage{subfigure}
\usepackage{algorithm}
\usepackage{algorithmic}
\usepackage{multirow}

\begin{document}

\title{Improving Resistance to Adversarial Deformations by Regularizing Gradients}

\author{Pengfei Xia}
\email{xpengfei@mail.ustc.edu.cn}

\author{Bin Li}
\authornote{Corresponding author.}
\email{binli@ustc.edu.cn}

\affiliation{
  \institution{University of Science and Technology of China}
  \city{Hefei}
  \state{China}
}

\begin{abstract}
Improving the resistance of deep neural networks against adversarial attacks is important for deploying models to realistic applications. However, most defense methods are designed to defend against intensity perturbations and ignore location perturbations, which should be equally important for deep model security. In this paper, we focus on adversarial deformations, a typical class of location perturbations, and propose a flow gradient regularization to improve the resistance of models. Theoretically, we prove that, compared with input gradient regularization, regularizing flow gradients is able to get a tighter bound. 

Over multiple datasets, architectures, and adversarial deformations, our empirical results indicate that models trained with flow gradients can acquire a better resistance than trained with input gradients with a large margin, and also better than adversarial training. Moreover, compared with directly training with adversarial deformations, our method can achieve better results in unseen attacks, and combining these two methods can improve the resistance further. 
\end{abstract}


\maketitle

\section{Introduction}
Deep neural networks (DNNs), especially convolutional neural networks (CNNs), have achieved remarkable success in computer vision tasks \cite{krizhevsky2012imagenet, simonyan2014very, girshick2015fast, long2015fully, xia2020boosting}. However, small, imperceptible changes to the underlying images can easily fool DNNs \cite{szegedy2013intriguing, goodfellow2014explaining}. Such modified inputs, also known as \textit{adversarial examples}, pose a doubt when applying deep learning models to security-sensitive applications, such as face recognition, surveillance, and self-driving cars \cite{shafahi2019adversarial, heaven2019deep}.

One of the primary principles of generating adversarial examples is that the instances with or without perturbations should look similar. Under this principle, the ways of performing such covert changes to 2D images can be roughly divided into two categories: intensity perturbations \cite{goodfellow2014explaining, madry2017towards, carlini2017towards} and location perturbations \cite{engstrom2017rotation, xiao2018spatially, alaifari2018adef}, where the former adds noise to each pixel and the latter modifies its position. Although these two types of attacks should be equally important for deploying DNNs to realistic applications, there are few works to study location perturbations, whether for attack or defense. The reason seems to be simple: location perturbations are not as convenient as intensity perturbations for theoretical analysis.

However, location perturbations cannot be ignored for deep learning security, because hackers will not have any burden to use any type of adversaries as long as they are offensive and concealed. Unfortunately, the performance of defense methods that designed for intensity perturbations, such as adversarial training \cite{goodfellow2014explaining, madry2017towards, wang2019bilateral}, feature squeezing \cite{xu2017feature}, and input gradient regularization \cite{ross2018improving, jakubovitz2018improving, chan2019jacobian}, cannot be guaranteed when against location perturbations \cite{engstrom2017rotation, xiao2018spatially}. How to defense location perturbations needs more attention.

In this paper, we focus on improving the resistance of models against adversarial deformations \cite{xiao2018spatially, alaifari2018adef}, a typical class of location perturbations. Specifically, adversarial deformations fool deep models by slightly flowing the position of each pixel in clean images, and restrict the flow to a small range to maintain high perceptual quality. To resist such attacks, an intuitive method is training model with adversarial deformations. However, the main drawbacks of training with adversarial examples are the high cost of training time \cite{zhang2019you, shafahi2019adversarial} and overfitting to the specific attack that appeared in training procedure \cite{tramer2019adversarial, tramer2020adaptive}. 

We suggest another method. The main principle we following here is that \textbf{the output of a robust model should be insensitive to a small variation of input}, where the key issue is how to measure the variation. Defensive methods adopting the same principle, such as input gradient regularization \cite{lyu2015unified, ross2018improving} and Jacobians regularization \cite{jakubovitz2018improving, hoffman2019robust}, use $l_p$-norm of the difference between the adversary and the clean image to measure the variation, which is designed for intensity perturbations and not suitable for measuring location perturbations. 

In order to effectively protect deep models from adversarial deformations, in this paper, we propose a flow gradient regularization that directly uses the degree of location flow to measure the variation. Theoretically, we prove that regularizing flow gradients is able to acquire a tighter bound than regularizing input gradients.

Our experimental results consistently show that training with flow gradients performs better than training with input gradients \cite{ross2018improving} with a large margin, and also better than adversarial training \cite{madry2017towards} when against adversarial deformations generated in four methods. Moreover, compared with training with adversarial deformations, flow gradient regularization performs better on attacks not seen in training, while these two methods can still be combined to improve resistance further.

\section{Related Works and Preliminaries}
\subsection{Adversarial Attacks}
Since Szegedy et al. \cite{szegedy2013intriguing} first noticed the existence of adversarial examples, many methods have been proposed for enhancing such attacks. Goodfellow et al. \cite{goodfellow2014explaining} provided a linear explanation of adversarial examples and proposed a single-step attack named fast gradient sign method (FGSM). Subsequently, some works have been done to expand it to multiple steps \cite{kurakin2016adversarial, dong2018boosting}. Among them, project gradient descent (PGD) proposed by Mądry et al. \cite{madry2017towards} is the most typical one and shows a strong attack ability. Carlini and Wagner \cite{carlini2017towards} proposed the famous C\&W attack, a powerful approach that regards generating adversarial examples as an optimization problem. Some works \cite{xiao2018generating, song2018constructing} construct an adversary by generative adversarial networks. Nowadays, building more aggressive and more concealed attacks is still a hot topic \cite{chen2018ead, ru2019bayesopt, croce2019minimally}.

Attacks methods mentioned above mostly fool deep models by adding a crafted noise to the clean image to perturb the intensity. Besides, there are some works performing chicaneries by shifting the location. Engstrom et al. \cite{engstrom2017rotation} found that convolutional neural networks are vulnerable to simple image transformations, such as rotation and translation. Xiao et al. \cite{xiao2018spatially} introduced adversarial deformations, which fool deep models by flowing the location in the input image, and limit the flow to a small range to keep visual similarity. Alaifai et al. \cite{alaifari2018adef} presented a method to find similar adversaries with a first-order optimizer. Zhang et al. \cite{zhang2019joint} combined both spatial and pixel perturbations and proposed a joint adversarial attack.

The diversity in generating methods for adversarial examples imposes great challenges to the research on constructing adversarially robust models.

\subsection{Adversarial Defenses}
To improve the resistance of models to such attacks, extensive efforts have come into the scene, such as preprocessing \cite{guo2017countering, xie2017mitigating, kou2019enhancing}, feature squeezing \cite{xu2017feature}, model ensemble \cite{tramer2017ensemble, sen2020empir} and certified defenses \cite{raghunathan2018certified}. The most direct and effective defenses so far are training models with generated adversarial examples as a kind of data augmentation. These adversarial training methods are first introduced by Goodfellow et al. \cite{goodfellow2014explaining} and developed by Mądry et al. \cite{madry2017towards}. Subsequently, research continued to be presented. Some works \cite{shafahi2019adversarial, zhang2019you} tried to decrease time consumption, which is one of the major drawbacks of adversarial training. Tramer and Boneh \cite{tramer2019adversarial} developed it to defense multiple attacks simultaneously.

The main idea of another type of defenses is to decrease the sensitivity of models' output to a small variation of input. Among them, the most typical ones are input gradient regularization \cite{lyu2015unified, ross2018improving} and Jacobians regularization \cite{jakubovitz2018improving}, where both are adding penalty items to the loss function during model training. Hoffman et al. \cite{hoffman2019robust} developed an efficient approximate algorithm to implement Jacobian regularizer. Chan et al. \cite{chan2019jacobian} proposed Jacobian adversarially regularized network to improve the saliency of Jacobians, and further increase robustness.

One concern is that these defenses mostly focus on intensity perturbations, and their resistance to location perturbations are less effective. Engstrom et al. \cite{engstrom2017rotation} showed that $l_{\infty}$-bounded adversarial training actually damages the accuracy of models to adversarial rotations and translations. Xiao et al. \cite{xiao2018spatially} tested different defenses against adversarial deformations and found that these methods can only achieve low defense performance. How to defense location perturbations is worthy of further research.

\section{Preliminaries}
\subsection{Intensity Perturbations}
Given a deep model $f$ and an input data pair $(x, y)$, the aim of intensity perturbations is to find a craft noise $\delta$ so that $f(x+\delta) \ne y$, where $x' = x + \delta$ denotes the generated adversary. To maintain visually imperceptible, $\|x'-x\|_p$ ($\|\delta\|_p$) is restricted to a small value. After determining the form of attack, researchers mainly focus on how to find a suitable $\delta$. In this paper, we review four types of methods, i.e., single-step attack, multi-step attack, optimization-based attack, and gradient-free attack, which are also used to generate adversarial deformations in our experiments.

Single-step attack (FGSM \cite{goodfellow2014explaining}) uses a single gradient ascent step to construct adversarial examples:
\begin{equation}
x' = x + \epsilon \cdot \operatorname{sign}(\nabla_{x}L(x, y)) \text{,}
\end{equation}
where $L$ denotes the loss function, $\operatorname{sign}$ denotes the sign function, and $\epsilon$ denotes a small value that specifying a noise budget.

Multi-step attack (BIM \cite{kurakin2016adversarial}, PGD \cite{madry2017towards}) is an extension of single-step that generating an adversary by iteratively calculating:
\begin{equation}
x^{k+1} = \operatorname{clip}(x^{k} + \alpha \cdot \operatorname{sign}(\nabla_{x^{k}}L(x, y))) \text{,}
\end{equation}
where $\alpha$ is the step size, $\operatorname{clip}$ is the clip function to ensure that $x^{k+1}$ is within a reasonable range, $x^{0} = x$, $x' = x^{K}$, and $K$ is the total number of iterations.

Optimization-based attack (L-BFGS \cite{szegedy2013intriguing}, C\&W \cite{carlini2017towards}) produces an adversarial for a target $t$ by minimizing the formulation:
\begin{equation}
L(x + \delta, t) + c \cdot \|\delta\|_p \text{,}
\end{equation}
where $c$ is a chosen constant that balancing the strength and the imperceptibility.

Unlike the above three attacks, which all use gradient information, gradient-free attack (ZOO \cite{chen2017zoo}, One Pixel \cite{su2019one}) only needs the classification confidence of models, and search $\delta$ by gradient estimation or evolutionary algorithm.

\subsection{Adversarial Deformations}
\begin{figure}[h]
\centering
\includegraphics[width=12.0cm]{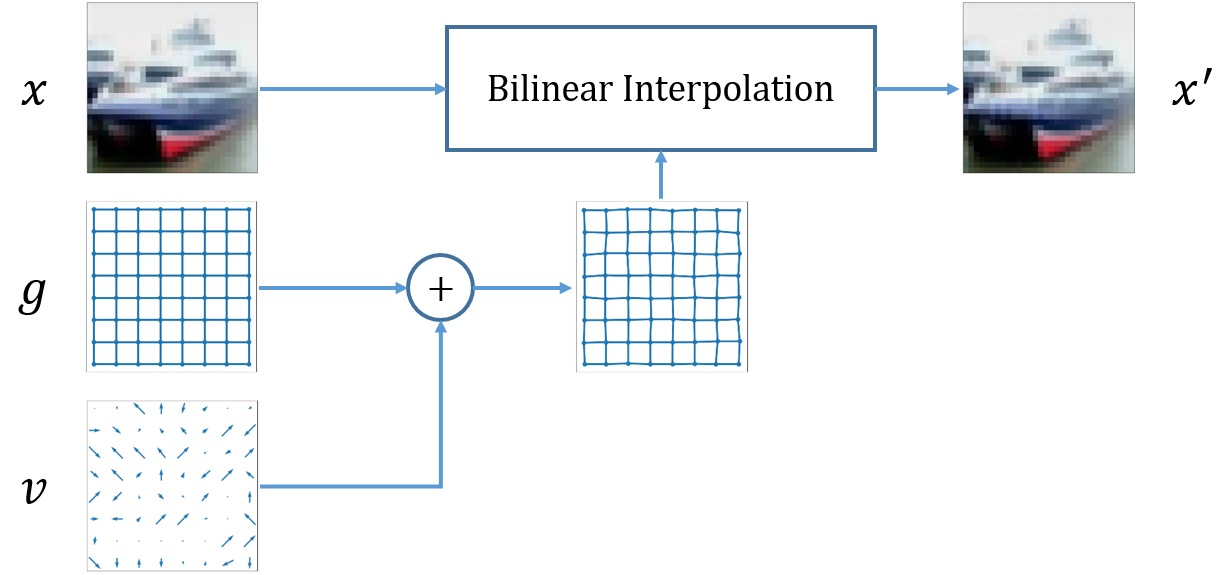}
\caption{The process of generating adversarial deformations.}
\label{fig:adversarial_deformations}
\end{figure}

As shown in Figure \ref{fig:adversarial_deformations}, we briefly introduce adversarial deformations, a typical class of location perturbations. Assuming the clean image $x \in \mathbb{R}^{W \times H \times C}$ with width $W$, height $H$ and channels $C$, $g = \{(m_i, n_i)\}_{i=1,...,W \times H} \in \mathbb{R}^{W \times H \times 2}$ is a 2D grid to denote the location of each pixel. Adversarial deformations fool deep models by defining a flow matrix $v$ to shift the location of each pixel, i.e., $v = \{(\Delta m_i, \Delta n_i)\}_{i=1,...,W \times H} \in \mathbb{R}^{W \times H \times 2}$. $I$ denotes the bilinear interpolation function, and the adversarial example is generated by $x' = I(x, g + v)$, and the $i$-th pixel of $x'$ is calculated as below:
\begin{equation}
x'_i = \sum_{q \in B(m_i', n_i')} x_q(1 - |m_i' - m_q|)(1 - |n_i' - n_q|) \text{,}
\end{equation}
where $m_i' = m_i + \Delta m_i$, $n_i' = n_i + \Delta n_i$ are the shifted location, $B(m_i', n_i')$ denotes the indices of 4-pixel neighbors at location $(m_i', n_i')$. In the following writing, we omit $g$ and use $x' = I(x, v)$ for simplicity.

Finding a feasible $v$ is similar to finding $\delta$ in intensity perturbations, and can be solved with the four aforementioned methods. For more details about adversarial deformations, please refer to \cite{xiao2018spatially, alaifari2018adef}.

\subsection{Input Gradient Regularization}
Input gradient regularization \cite{ross2018improving} influences the training process by adding $\|\nabla_{x} L(x, y)\|_p$ as the penalty item to reduce the sensitivity of output to the input variation, that is:
\begin{equation}
Loss = L(x, y) + \lambda \cdot \|\nabla_{x} L(x, y)\|_p \text{,}
\end{equation}
where $\lambda$ denotes a hyperparameter that controls the penalty strength.

\subsection{Adversarial Training}
Adversarial training improves robustness by optimizing:
\begin{equation}
\mathop{\arg\min}_{\theta} \mathbb{E}_{(x, y)} \big[ \max_{\|\delta\|_p \le \epsilon} L_{\theta}(x + \delta, y) \big] \text{,}
\end{equation}
where $\theta$ denotes the parameter of the model. It should be note that, in this paper, adversarial training specifically refers to training with intensity perturbation. We call training with adversarial deformations as adversarial deformation training.

\section{Methodology}
\subsection{Flow Gradient Regularization}
For a robust classifier, the output should be insensitive to the small variation of input \cite{lyu2015unified, simon2019first}. Many methods have been proposed to achieve the goal, like regularizing the Frobenius norm of the Jacobian matrix of models evaluated on the output data \cite{jakubovitz2018improving, hoffman2019robust}. However, the $l_p$-norm distance in input space, i.e., $\|x' - x\|_p$, used in these methods is not entirely suitable for measuring adversarial deformations. This paper shows a method that regularizing the gradient of loss to the flow matrix.

Defining $L(x, y)$ as a loss of the model $f$, and $x$, $y$ are the input image and its ground-truth label respectively. Suppose $\Delta L$ is a variation of the loss, that is:
\begin{equation}
\Delta L = |L(x', y) - L(x, y)| \text{.}
\label{equ:variation1}
\end{equation}
If taking $\|x'-x\|_p$ as the variation metric, we can directly perform Taylor expansion on $L(x', y)$ around point $x$ and derive the form of input gradient regularization \cite{ross2018improving}. Instead, we use $\|v - v_0\|_p$ in this paper, where $v_0$ is the identity value so that $x = I(x, v_0)$. Substituting $x' = I(x, v)$ and $x = I(x, v_0)$ into Equation \ref{equ:variation1}, we can get:
\begin{equation}
\Delta L = |L(I(x, v), y) - L(I(x, v_0), y)| \text{.}
\label{equ:variation2}
\end{equation}

Approximating the loss function around $v_0$ by the first-order Taylor expansion and ignoring the higher order terms, we can get:
\begin{equation}
L(I(x, v), y) = L(I(x, v_0), y) + \nabla_{v_0}L \cdot (v - v_0) \text{,}
\end{equation}
where $\nabla_{v_0}L$ is a short form for $\nabla_{v}L(I(x, v), y)|_{v = v_0}$.

Substituting it into Equation \ref{equ:variation2} gives:
\begin{equation}
\Delta L = |\nabla_{v_0} L \cdot (v - v_0)| \le \|\nabla_{v_0} L\|_q \cdot \|v - v_0\|_p,
\label{equ:variation3}
\end{equation}
where the Hölder inequality is used and $\frac{1}{p} + \frac{1}{q} = 1$.

The proposed method reduces the impact of input disturbance on the output by adding the flow gradients $\|\nabla_{v_0} L\|_q$ as a penalty item to the loss function, and the total loss is
\begin{equation}
Loss = L(x, y) + \lambda \cdot \|\nabla_{v_0} L(I(x, v), y)\|_q^2 \text{,}
\end{equation}
where $\lambda$ is a hyperparameter specifying the penalty strength, and we set $p=q=2$ in the following experiments.

\subsection{Theoretical Analysis}
Next we explore the relationship between input gradients $\|\nabla_{x} L(x, y)\|_q$ and flow gradients $\|\nabla_{v_0} L\|_q$. According to the chain rule, $\|\nabla_{v_0} L\|_q$ can be written as:
\begin{equation}
\|\nabla_{v_0} L\|_q = \|\nabla_{x'} L(x', y) \cdot \nabla_{v_0} I(x, v)\|_q \text{.}
\end{equation}
Since $\nabla_{x'} L(x', y)$ is equivalent to $\nabla_{x} L(x, y)$ in input gradient regularization, we can get:
\begin{equation}
\|\nabla_{v_0} L\|_q = \|\nabla_{x} L(x, y) \cdot \nabla_{v_0} I(x, v)\|_q \le \|\nabla_{x} L(x, y)\|_q \cdot \|\nabla_{v_0} I(x, v)\|_q \text{.}
\end{equation}

Considering the derivative part, $I$ satisfies Lipschitz constraint, that is, the norm of the derivative of $I$ is bounded. Let
\begin{equation}
C = \sup_x \|\nabla_{v_0} I(x, v)\|_q \text{,}
\end{equation}
then
\begin{equation}
\|\nabla_{v_0} L\|_q \le C \cdot \|\nabla_{x} L(x, y)\|_q \text{.}
\end{equation}

Substitute it into Equation \ref{equ:variation3}, we can get:
\begin{equation}
\Delta L \le \|\nabla_{v_0} L\|_q \cdot \|v - v_0\|_p \le C \cdot \|\nabla_{x} L(x, y)\|_q \cdot \|v - v_0\|_p \text{,}
\end{equation}
which indicates that, for adversarial deformations, $\|\nabla_{v_0} L\|_q$ is a tighter bound than $\|\nabla_{x} L(x, y)\|_q$. Therefore, theoretically, regularizing flow gradients can better resist adversarial deformations than regularizing input gradients.

\section{Experiments}
\subsection{Setup}
We conduct experiments on CIFAR-10 and CIFAR-100 \cite{krizhevsky2010convolutional} with VGG-11 \cite{simonyan2014very} and ResNet-18 \cite{he2016deep}. For all experiments, we use SGD optimizer with momentum of 0.9 and weight decay of 5e-4. The batch size is set to 0.9 and the total training duration is 60 epochs. The initial learning rate is 0.1 and is dropped by 10 after 30 and 50 epochs. We employ random cropping and random flipping as data augmentations. All experiments are implemented with PyTorch \cite{paszke2017automatic} and run on a GeForce GTX 2080 Ti.

\subsection{Attacks}
All trained models are evaluated against adversarial deformations generated by four methods: single-step attack, multi-step attack, optimization-based attack, and gradient-free attack. Single-step and multi-step attacks are extending FGSM \cite{goodfellow2014explaining} and PGD \cite{madry2017towards} respectively. For the optimization-based attack, we adopt stAdv \cite{xiao2018spatially} and solve the problem with SGD. Evolution strategies \cite{rechenberg1978evolutionsstrategien} is used in gradient-free attack to find a suitable $v$ that fools DNNs.

\subsection{Defenses}
To test the proposed methods, the following defenses are used to train a deep model:
\begin{itemize}
\item Standard, training without using any defense methods;
\item Adversarial training (AT), training with intensity perturbations generated by $l_{\infty}$-bounded PGD \cite{madry2017towards}, and setting $\epsilon$ to $8 / 255$, $K$ to 7.
\item Input gradient regularization (IGR), training with input gradients as a penalty \cite{ross2018improving}, and setting $\lambda$ to $\{3000, 5000, 7000\}$ for all datasets;
\item Adversarial deformation training (ADT), training with adversarial deformations generated by multi-step attack, and setting the flow budget to 0.01, the number of iterations to 7; 
\item Flow gradient regularization (FGR), training with flow gradients as a penalty, and setting $\lambda$ to $\{400, 700, 1000\}$ for all datasets;
\item Flow gradient regularization with adversarial deformation training (FGR+ADT), a combination of these two methods.
\end{itemize}
It should be noted that the selection of the above parameters, one part is to refer to the previous literature \cite{madry2017towards}, such as 7 iterations for AT, and the other part is to ensure that all methods can acquire similar accuracy on clean images for a fair comparison.

\subsection{Results}
\begin{figure}[]
\centering
\subfigure[CIFAR-10, VGG-11, MS] {\includegraphics[width=6.5cm]{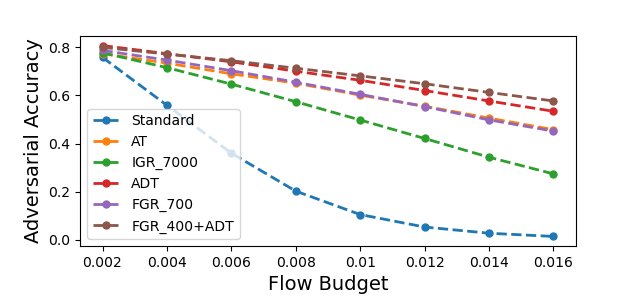}}
\subfigure[CIFAR-10, VGG-11, OB] {\includegraphics[width=6.5cm]{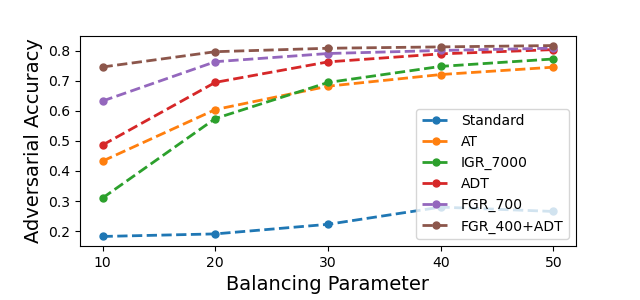}}

\subfigure[CIFAR-10, ResNet-18, MS] {\includegraphics[width=6.5cm]{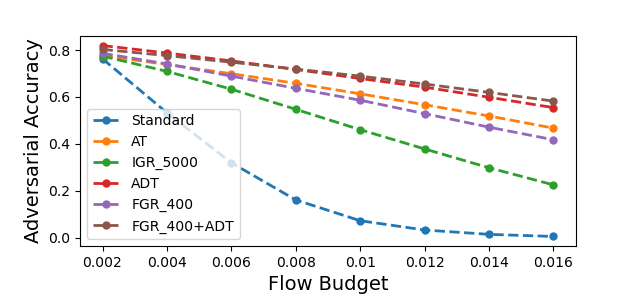}}
\subfigure[CIFAR-10, ResNet-18, OB] {\includegraphics[width=6.5cm]{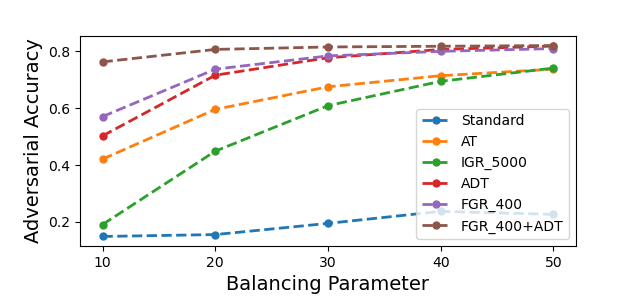}}

\subfigure[CIFAR-100, VGG-11, MS] {\includegraphics[width=6.5cm]{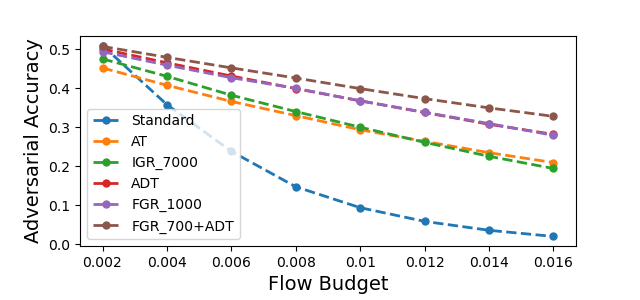}}
\subfigure[CIFAR-100, VGG-11, OB] {\includegraphics[width=6.5cm]{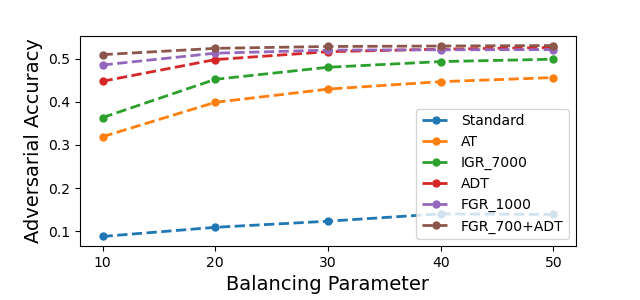}}

\subfigure[CIFAR-100, ResNet-18, MS] {\includegraphics[width=6.5cm]{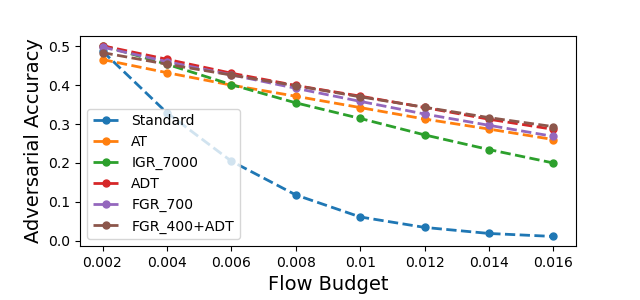}}
\subfigure[CIFAR-100, ResNet-18, OB] {\includegraphics[width=6.5cm]{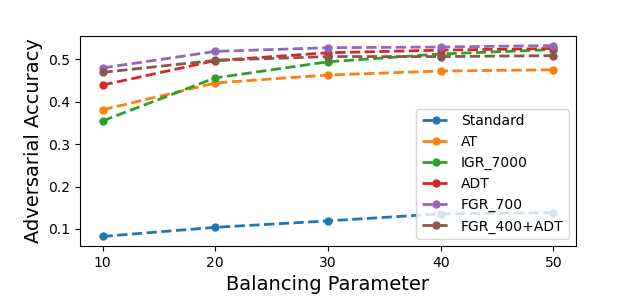}}

\caption{Performances of defenses in resisting multi-step attack (MS) with different flow budget $\epsilon$ and optimization-based attack (OB) with different balancing parameter $c$.}
\label{fig:cifar_results}
\end{figure}

Our experimental results on CIFAR-10 and CIFAR-100 against different adversarial deformations are shown in Table \ref{tab:cifar_results} and Figure \ref{fig:cifar_results}. Some observations are summarized as follows.

\begin{table}[h] 
\footnotesize
\centering 
\caption{Results on CIFAR-10 and CIFAR-100 against adversarial deformations. V-11: VGG-11. R-18: ResNet-18. SS: single-step attack with the flow budget is 0.01. MS: multi-step attack with the flow budget is 0.01 and the total number of iterations is 20. OB: optimization-based attack with SGD optimizer and balancing parameter $c$ is 10. GF: gradient-free attack with ES algorithm and $c$ is 10.}
\begin{tabular}{l|c|c|c|c|c|c|c|c|c|c} 
\toprule
\multirow{2}{*}{Defense}       & \multicolumn{5}{c|}{CIFAR-10}         & \multicolumn{5}{c}{CIFAR-100}         \\
                               & Clean & SS    & MS    & OB    & GF    & Clean & SS    & MS    & OB    & GF    \\ \midrule
V-11, Standard                 & \textbf{0.901} & 0.385 & 0.105 & 0.182 & 0.062 & \textbf{0.669} & 0.246 & 0.094 & 0.088 & 0.032 \\
V-11, AT                       & 0.808 & 0.646 & 0.601 & 0.433 & 0.423 & 0.495 & 0.327 & 0.293 & 0.319 & 0.189 \\
V-11, IGR, $\lambda=3000$      & 0.851 & 0.541 & 0.432 & 0.136 & 0.228 & 0.581 & 0.340 & 0.282 & 0.262 & 0.074 \\
V-11, IGR, $\lambda=5000$      & 0.840 & 0.562 & 0.476 & 0.211 & 0.237 & 0.556 & 0.342 & 0.299 & 0.331 & 0.100 \\
V-11, IGR, $\lambda=7000$      & 0.829 & 0.571 & 0.498 & 0.310 & 0.276 & 0.518 & 0.337 & 0.300 & 0.363 & 0.115 \\
V-11, ADT                      & 0.828 & 0.698 & 0.663 & 0.486 & 0.577 & 0.538 & 0.398 & 0.367 & 0.448 & 0.323 \\
V-11, FGR, $\lambda=400$       & 0.844 & 0.639 & 0.586 & 0.504 & 0.535 & 0.575 & 0.393 & 0.355 & 0.456 & 0.323 \\
V-11, FGR, $\lambda=700$       & 0.825 & 0.649 & 0.604 & 0.633 & 0.571 & 0.551 & 0.398 & 0.369 & 0.494 & \textbf{0.372} \\
V-11, FGR, $\lambda=1000$      & 0.807 & 0.641 & 0.605 & 0.689 & 0.589 & 0.525 & 0.392 & 0.367 & 0.485 & 0.363 \\
V-11, FGR+ADT, $\lambda=400$   & 0.827 & \textbf{0.706} & \textbf{0.681} & 0.746 & 0.617 & 0.541 & 0.407 & 0.381 & 0.489 & 0.352 \\
V-11, FGR+ADT, $\lambda=700$   & 0.817 & 0.704 & \textbf{0.681} & \textbf{0.765} & \textbf{0.632} & 0.535 & \textbf{0.420} & 0.399 & 0.509 & 0.352 \\
V-11, FGR+ADT, $\lambda=1000$  & 0.801 & 0.686 & 0.663 & 0.759 & 0.629 & 0.525 & \textbf{0.420} & \textbf{0.402} & \textbf{0.509} & \textbf{0.372} \\ \midrule
R-18, Standard                 & \textbf{0.910} & 0.363 & 0.072 & 0.149 & 0.058 & \textbf{0.663} & 0.222 & 0.061 & 0.082 & 0.028 \\
R-18, AT                       & 0.807 & 0.655 & 0.613 & 0.421 & 0.357 & 0.498 & 0.368 & 0.343 & 0.381 & 0.206 \\
R-18, IGR, $\lambda=3000$      & 0.845 & 0.541 & 0.437 & 0.114 & 0.198 & 0.597 & 0.347 & 0.280 & 0.236 & 0.069 \\
R-18, IGR, $\lambda=5000$      & 0.831 & 0.549 & 0.460 & 0.191 & 0.241 & 0.571 & 0.359 & 0.307 & 0.302 & 0.077 \\
R-18, IGR, $\lambda=7000$      & 0.818 & 0.562 & 0.485 & 0.278 & 0.248 & 0.548 & 0.355 & 0.315 & 0.355 & 0.087 \\
R-18, ADT                      & 0.837 & \textbf{0.720} & 0.679 & 0.503 & 0.557 & 0.539 & \textbf{0.401} & 0.372 & 0.439 & 0.320 \\
R-18, FGR, $\lambda=400$       & 0.830 & 0.633 & 0.586 & 0.570 & 0.542 & 0.559 & 0.389 & 0.353 & 0.447 & 0.346 \\
R-18, FGR, $\lambda=700$       & 0.804 & 0.637 & 0.596 & 0.669 & 0.579 & 0.538 & 0.391 & 0.359 & 0.480 & 0.354 \\
R-18, FGR, $\lambda=1000$      & 0.790 & 0.638 & 0.602 & 0.691 & 0.584 & 0.509 & 0.382 & 0.355 & 0.481 & 0.367 \\
R-18, FGR+ADT, $\lambda=400$   & 0.828 & 0.716 & \textbf{0.689} & \textbf{0.763} & \textbf{0.647} & 0.517 & 0.396 & 0.370 & 0.469 & 0.334 \\
R-18, FGR+ADT, $\lambda=700$   & 0.802 & 0.694 & 0.672 & 0.758 & 0.640 & 0.509 & 0.394 & 0.374 & 0.476 & 0.363 \\
R-18, FGR+ADT, $\lambda=1000$  & 0.786 & 0.683 & 0.659 & 0.751 & 0.635 & 0.503 & 0.399 & \textbf{0.379} & \textbf{0.485} & \textbf{0.367} \\
\bottomrule
\end{tabular}
\label{tab:cifar_results}
\end{table}

\noindent{\textbf{FGR versus IGR.}} With similar accuracy on the clean set, training with FGR is significantly better than training with IGR, which also empirically confirms the above analysis of constraining the flow gradients can get a tighter bound. Specifically, our results indicate, compared with IGR, FGR can bring additional adversarial accuracy improvements in all cases, where the average improvements for single-step, multi-step, optimization-based and gradient-free attacks are 8.5\%, 13.2\%, 41.9\%, 32.9\% for CIFAR-10, and 4.4\%, 6.2\%, 16.6\%, 26.7\% for CIFAR-100. Relatively speaking, FGR improves more on optimization-based attack and gradient-free attack.

\noindent{\textbf{FGR versus AT.}} We choose a suitable $\lambda$ for FGR that leads to similar accuracy with AT on clean set for a fair comparison about these two methods. The model trained with FGR is leading in most cases, especially in optimization-based attack and gradient-free attack. Both the performances of IGR and AT support the view that defenses customized for intensity perturbations are not suitable for location perturbations.

\noindent{\textbf{FGR versus ADT.}} FGR and ADT have their own advantages. In general, adversarial deformation training is better at resisting single-step and multi-step attacks, and models trained with FGR get stronger resistance to optimization-based and gradient-free attacks. We argue this is because in ADT, samples are generated through multi-step method, so trained models will be somewhat ``overfitting'' to the iterative attacks. Relatively, FGR belongs to gradient regularization and does not need to know in advance what method to attack.

\noindent{\textbf{Combining FGR with ADT.}} In general, the combination of FGR and ADT can bring the best resistance to deep models against adversarial deformations generated with four methods.

\subsection{Loss Landscape}
\begin{figure}[]
\centering
\subfigure[Standard] {\includegraphics[width=6.5cm]{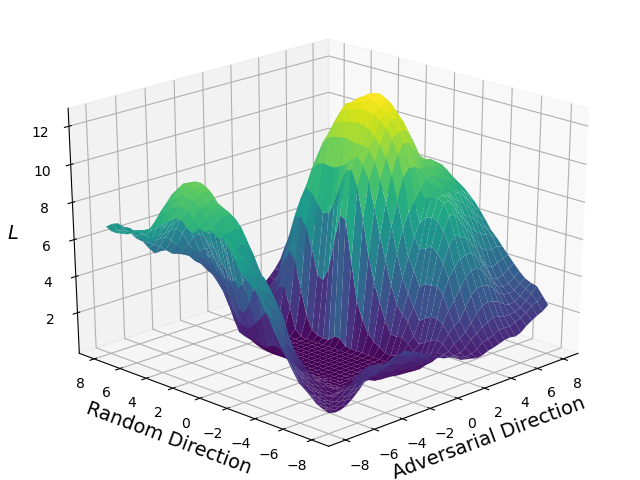}}
\subfigure[AT] {\includegraphics[width=6.5cm]{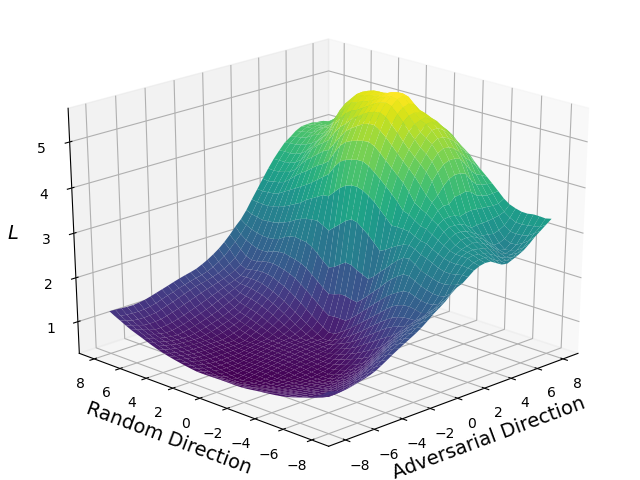}}
\subfigure[IGR, $\lambda=7000$] {\includegraphics[width=6.5cm]{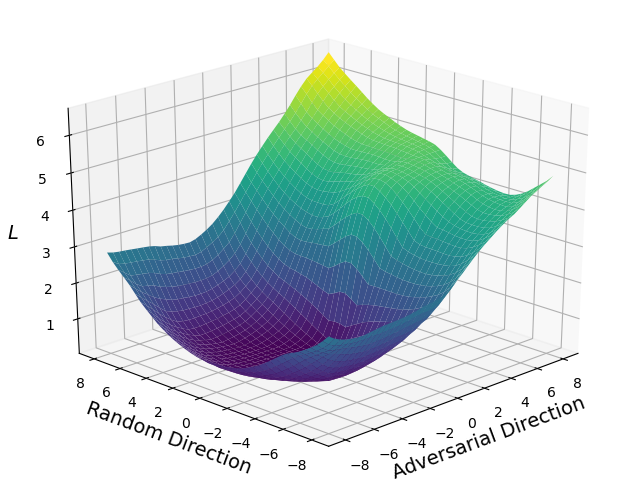}}
\subfigure[ADT] {\includegraphics[width=6.5cm]{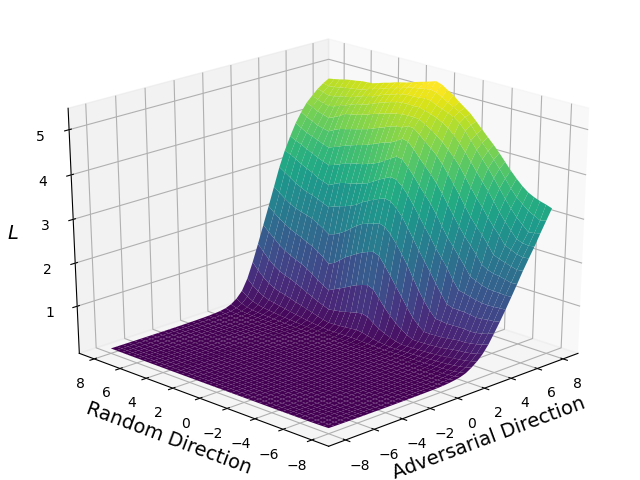}}
\subfigure[FGR, $\lambda=400$] {\includegraphics[width=6.5cm]{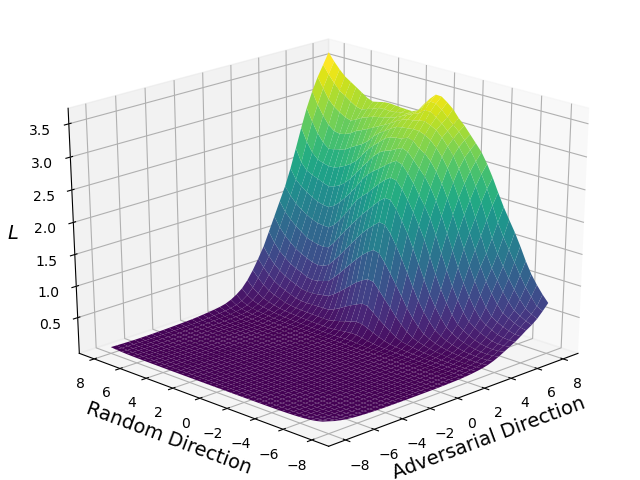}}
\subfigure[FGR, $\lambda=400$ + ADT] {\includegraphics[width=6.5cm]{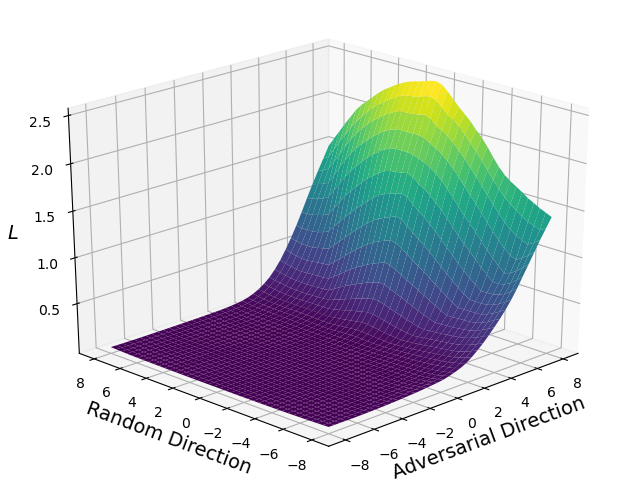}}

\caption{Loss surfaces along the adversarial flow and a random flow on CIFAR-10 with VGG-11.}
\label{fig:cifar_loss_landscape}
\end{figure}

The classification loss values are computed along the adversarial flow direction and a random flow direction to analyze the loss landscape of the models trained with different defense methods, as shown in Figure \ref{fig:cifar_loss_landscape}.

Compared with standard training, both IGR, AT, ADT and FGR can low the amplitudes. It indicates that all defenses can reduce the sensitivity of models' output to a small variation of input, and is consistent with the results in Table \ref{tab:cifar_results} and Figure \ref{fig:cifar_results}.

However, the landscape changes in these defenses behave differently. First of all, the landscapes of defenses designed for intensity perturbations, including IGR and AT, are not very flat, especially away from the center. The main reason for this is that $\|x'-x\|_p$ is not entirely suitable for measuring the variation generated by adversarial deformations. Secondly, compared with ADT, FGR has a lower amplitude. In general, combining FGR and ADT can achieve the best results in both smoothness and amplitude.

\section{Conclusion}
In order to increase the resistance of deep neural networks against adversarial deformations, a typical type of location perturbations, we propose a defense method named flow gradient regularization, which adds the flow gradients as a penalty to the loss function. The proposed defense is evaluated on CIFAR-10 and CIFAR-100 against adversarial deformations generated with four methods, i.e., single-step attack, multi-step attack, optimization-based attack and gradient-free attack. The results consistently show that, compared with IGR and AT, models trained with FGR can get better resistance with a large margin. The comparison with ADT indicates that FGR is more suitable for resisting unseen attacks. Moreover, these two methods can be combined to improve models' robustness further. The visualization of loss landscapes also supports these conclusions.

\bibliographystyle{ACM-Reference-Format}
\bibliography{./references}

\end{document}